\documentclass[11pt]{article} 
\usepackage{rldmsubmit,palatino}
\usepackage{graphicx}
\usepackage{subfigure}
\usepackage{amsmath,amsfonts}
\usepackage{algorithmic}
\usepackage{algorithm}
\usepackage{array}
\usepackage{graphicx}
\usepackage{multirow}
\DeclareMathOperator{\E}{\mathbb{E}}

\title{A Motivational Architecture for Open-Ended Learning Challenges in Robots}

\author{
Alejandro Romero $^{1}$, Gianluca Baldassarre $^{2}$, Richard J. Duro$^{1}$, Vieri Giuliano Santucci$^{2}$\\
$^1$Integrated Group for Engineering Research (GII)\\
CITIC research center\\
Universidade da Coruña, Spain\\
\texttt{\{alejandro.romero.montero, richard.duro\}@udc.es}\\
$^2$Istituto di Scienze
e Tecnologie della Cognizione (ISTC)\\
Consiglio Nazionale delle Ricerche (CNR), Roma, Italy\\
\texttt{\{gianluca.baldassarre, vieri.santucci\}@istc.cnr.it}\\
}

\begin{document}

\maketitle

\begin{abstract}
Developing agents capable of autonomously interacting with complex and dynamic environments, where task structures may change over time and prior knowledge cannot be relied upon, is a key prerequisite for deploying artificial systems in real-world settings. The open-ended learning framework identifies the core challenges for creating such agents, including the ability to autonomously generate new goals, acquire the necessary skills (or curricula of skills) to achieve them, and adapt to non-stationary environments. While many existing works tackles various aspects of these challenges in isolation, few propose integrated solutions that address them simultaneously. In this paper, we introduce H-GRAIL, a hierarchical architecture that, through the use of different typologies of intrinsic motivations and interconnected learning mechanisms, autonomously discovers new goals, learns the required skills for their achievement, generates skill sequences for tackling interdependent tasks, and adapts to non-stationary environments. We tested H-GRAIL in a real robotic scenario, demonstrating how the proposed solutions effectively address the various challenges of open-ended learning.
\end{abstract}


\acknowledgements{This work was partially funded by MCIN/AEI/10.13039/501100011033 (PID2021-126220OB-I00),
by ``ERDF A way of making Europe'', Xunta de Galicia (EDC431C-2021/39), Centro de Investigación de Galicia ``CITIC'' (ED431G 2019/01), and by the European Union’s Horizon 2020, Research and Innovation Programme, GA 101070381 (``PILLAR-Robots - Purposeful Intrinsically-motivated Lifelong Learning Autonomous Robots''), and NextGenerationEU PNRR, MUR code PE0000013, CUP B53C22003630006 (project ``FAIR - Future Artificial Intelligence Research'').}

\startmain 

\section{Introduction}
\label{sec:intro}

Recent AI and robotics advancements suggest a future where artificial agents are widely employed in human activities. However, in unstructured or unknown environments, pre-training is insufficient, as task requirements are often unpredictable. Most agents still rely on human-designed behaviors, lacking the autonomy to handle complex, dynamic scenarios. 
Unlike continual or lifelong learning, open-ended learning (OEL) \cite{Sigaud2023OELdefinition} emphasizes autonomous goal discovery, inherently tied to challenges like competence acquisition, curriculum learning, and adaptation in non-stationary environments. This integration requires agents to identify, prioritize, and learn interdependent tasks while addressing evolving goals and scenarios, demanding approaches that go beyond solving these challenges in isolation.

Research within Developmental Robotics and machine learning has advanced solutions to specific OEL challenges. Intrinsically motivated open-ended learning (IMOL) \cite{Santucci2020intrinsically} fosters exploration and competence acquisition without predefined tasks, utilizing novelty and information-theoretic metrics to enhance exploration \cite{Pathak2017curiosity}. Other strategies focus on developing autonomous curricula of increasingly complex tasks  
and adaptive knowledge representations \cite{Colas2023Augmenting}. However, despite the progress made, most approaches remain limited by their focus on isolated issues, preventing them from addressing multiple challenges simultaneously. True autonomous OEL agents require integrated systems capable of handling various aspects together \cite{LeCun2022path}. While some existing architectures attempt to combine mechanisms for different OEL components \cite{blaes2019control,forestier2022}, they still fall short in addressing the full range of OEL challenges.

To advance towards integrated architectures for autonomous intrinsically-motivated OEL, we propose "H-GRAIL - Hierarchical Goal-discovering Robotic Architecture for Intrinsically-motivated Learning," a cognitive architecture that enhances the GRAIL framework \cite{Santucci2016grail,Romero2022ICDL_HGRAIL}. H-GRAIL enables an agent to autonomously tackle goal discovery, competence acquisition, and manage intrinsic motivations while handling interdependent and non-stationary task scenarios. This architecture integrates these functionalities into a unified system, addressing the complex interdependence between goal discovery and other OEL challenges, requiring novel solutions beyond isolated issues.


\section{Problem Analysis and Proposed Solutions}
\label{sec:Overview}

From the perspective of the OEL framework, the general problem we are tackling is that of an agent that must autonomously learn to interact with an environment that may be unknown at design time, acquiring competences that may serve as preconditions for others, with transitions potentially changing over time. This is the process of gaining cumulative knowledge to achieve specific target states (goals) $G \in \mathcal{G}$, where each $G$ is a subset of states $G \subset S$ and it is achieved when $s \in G$ is reached. While some works \cite{Florensa2018,forestier2022} assume the agent knows the set of all goals $\mathcal{G}$, we consider the case where the agent initially lacks knowledge of $\mathcal{G}$. In this case, the known goals set $\mathcal{G}_k \subseteq \mathcal{G}$ is empty, and the agent's task is to continuously discover new goals $G$ to guide the acquisition of new competences (we will discuss this later).

Given $\mathcal{G}_k$, the agent's goal is to acquire competence in achieving all $G \in \mathcal{G}_k$. This can be framed as a multi-task reinforcement learning problem \cite{Florensa2018}, where the agent trains policies $\pi^G$, within a finite and unknown time, to maximise the expected overall competence $C =\E C^G$, where $C^G$ is the competence in achieving $G$. This problem can be seen as an N-armed bandit (with $N = |\mathcal{G}_k|$), where at each step $t$, the agent selects a goal $G$ to maximise $C$ over time. The agent evaluates each goal based on its potential competence gain, which, as shown in the literature, enables fast autonomous learning in robotic environments \cite{Santucci2013best}. In particular, the competence improvement $\Delta C^G$ after training on $G$ for time $h$ is given by:  $ \Delta C^G = C^G(t+h) - C^G(t)$.  Once $G$ is selected, the related policy $\pi^G$ can be trained through any algorithm optimising the achievement of $G$. 

Thus, learning multiple goals in OEL involves a high-level task of goal selection and a low-level task of policy training, but this problem needs to be reformulated when goals are interdependent, with dependencies potentially changing over time. In this setting, a goal $G$ can only be obtained if a subset of other goals, $\mathcal{G}_p^G \subseteq \mathcal{G} _k$ (the ``preconditions''), has already been achieved. Training low-level policies $\pi^G$ ensuring that all precondition for $G$ are met can be inefficient as the number of preconditions increases \cite{Romero2021ICAR}. A more efficient approach is to combine modular policies into curricula of subsequent skills matching all the preconditions for $G$. While learning a world model could help combine interrelated skills \cite{blaes2019control}, in a model-free approach, the solution is to learn a meta-policy $\Pi^G$ that selects the correct sequence of (sub-) goals $\mathcal{S}_G =\ <G_{s1},..., G_{si}>$, thus the related policies $\pi^{G_i}$, constituting the preconditions for $G$. (Note that each goal $G \in \mathcal{G}_k$ can also be a sub-goal $G_{si} \in \mathcal{S}_{G'}$ of another goal $G'$, and we will use the latter notation for simplicity). This introduces a Markov Decision Process (MDP) where, given the selected goal $G$ and the current state $s_t$, the meta-policy $\Pi^G$ learns to select the proper sequence of sub-goals to maximize the achievement of $G$.
Having the ability to choose the appropriate sub-goal based on the current state prevents wasting time on goals whose preconditions haven't been met \cite{forestier2022}. Furthermore, although low-level skills in interdependent goal scenarios could be learned by collapsing goal and sub-goal selection into a single layer, this would constrain the ability to train and store curricula needed to sequence skills that serve as preconditions for different goals \cite{Romero2022ICDL_HGRAIL}. Finally, in non-stationary scenarios where goal dependencies change over time, the meta-policies $\Pi^{G}$ must include dynamic adaptability mechanisms: the estimation of the current competence over a goal $C^G$ provides a self-generated modulation of the activity of $\Pi^{G}$ in non-stationary scenarios \cite{Santucci2016grail,Romero2022ICDL_HGRAIL}.

Up to this point, we have addressed the problem of autonomously selecting target goals, learning sub-goal sequences for interdependent tasks, and developing goal-specific policies. In OEL, however, the system must also discover new goals, so that initially $\mathcal{G}_k = \emptyset$. In literature, goal discovery is often a byproduct of skill learning \cite{Santucci2016grail}, 
where the agent encounters ``interesting states'' during exploration: while untrained policies may discover new goals, this process becomes inefficient once policies start fitting goal achievement. Other works pre-learn goals via exploratory strategies (e.g., \cite{Eysenbach2019diversity})
, but fail to leverage the skills gradually acquired to scaffold the discovery of more complex goals in interdependent scenarios. To overcome these challenges, we propose a twofold solution: goal discovery as an intrinsic motivation that the system can autonomously select, and the use of acquired competencies to guide goal discovery. 

\begin{figure}
    \centering
    \includegraphics[width=0.36\textwidth]{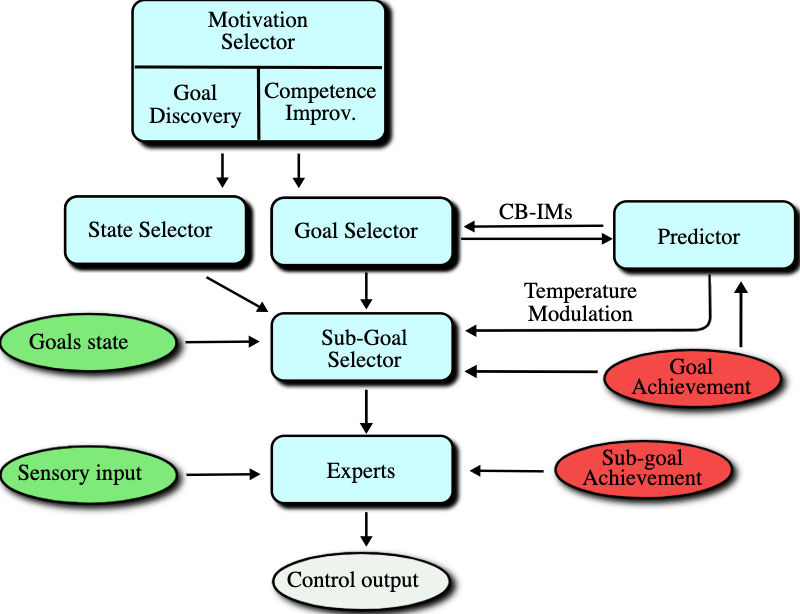}
   \caption{H-GRAIL, the proposed robotic architecture.}
    \label{Fig:Architecture}
\end{figure}

The way we address the different problems related to OEL, leads to the following hierarchical architecture (Fig. \ref{Fig:Architecture}): (1) a \textit{Motivation Selector} to decide between goal discovery and competence maximisation; (2) a \textit{Goal Selector} to choose a goal (or starting state for exploration); (3) a \textit{Sub-Goal Selector} to learn the sequence of sub-goals for the selected goal; and (4) \textit{low-level policies} to control robot movements and achieve the (sub-)goals.

\section{H-GRAIL}
\label{sec:solution}
This section details the implementation of the H-GRAIL architecture. The system, shown in Fig. \ref{Fig:Architecture}, integrates key mechanisms for motivation, goal selection, learning, and adaptation, explaining how they can be implemented in practical experiments while allowing flexibility for alternative algorithmic solutions.

The \textbf{Motivation Selector} employs an N-armed bandit to alternate between two intrinsic motivations: novelty-driven goal discovery and competence improvement, with a softmax rule guiding selection. Competence improvement is dynamically adjusted based on the highest observed competence gain across discovered goals.

The \textbf{Goal Selector}, also implemented as an N-armed bandit, prioritizes specific goals using competence-based intrinsic motivations (CB-IMs). These are calculated via an exponential moving average of competence improvements over successive trials.

The \textbf{Sub-Goal Selector} utilizes Q-Learning to decompose higher-level goals into actionable sub-goals, adjusting its exploration-exploitation balance to cope with environmental non-stationarity. The exploratory behavior is informed by the agent's current competence in achieving the primary goal.

The \textbf{Experts} module learns policies for achieving goals using neural network-based utility models. Experts evaluate candidate actions and select those maximizing the probability of success for a given goal.

The \textbf{State Selector}, inspired by Go-Explore \cite{Ecoffet2019GoExplore}, autonomously determines initial states for exploration based on historical success in discovering new goals from specific starting states. The robot then configures the environment accordingly, leveraging exploratory experts to identify novel states.

Finally, for \textbf{goal discovery and representation}, the system identifies ``interesting'' states via changes in visual sensory input, storing them as goal representations in a Goal Representation Map (GR-M). This map associates goal images with selector units, enabling recognition of goal achievement. In addition, there is a \textbf{Goal Matching} mechanism that autonomously evaluates whether goals are achieved by comparing the current sensory input against stored goal representations, providing binary feedback to reinforce learning and guide future goal selection.


\section{Experimental Setup}
\label{sec:ExpSetup}

To illustrate the functionality of the proposed architecture, we developed a robotic scenario designed for open-ended learning. In these scenario, the robot is tasked with (1) discovering potential goals, (2) acquiring the skills needed to achieve them, (3) learning the interdependencies among these goals, and (4) adapting to changes involving interdependencies that shift over time and goals that emerge or vanish during the experiment. Additionally, we compare the results obtained with H-GRAIL to those of several benchmark systems present in the literature:
\begin{itemize}
    \item \textit{Rnd-GD} (Random Goal Discovery): A system similar to H-GRAIL but without a Motivation Selector, relying on random or incidental goal discovery.  This approach reflects methods in the literature \cite{Santucci2016grail, forestier2022} where goal discovery emerges as a byproduct of other processes, such as competence improvement or reward maximization.
    \item \textit{S-GD} (Simple Goal Discovery): A system with a Motivation Selector but lacking a State Selector, allowing explicit but unguided exploration without the ability to strategically select starting points. This setup mirrors many systems \cite{romero2019bootstrapping,Eysenbach2019diversity} that separate goal discovery and exploitation phases.
\end{itemize}

\begin{figure}
    \centering
    \subfigure[]{\includegraphics[width=0.25\textwidth]{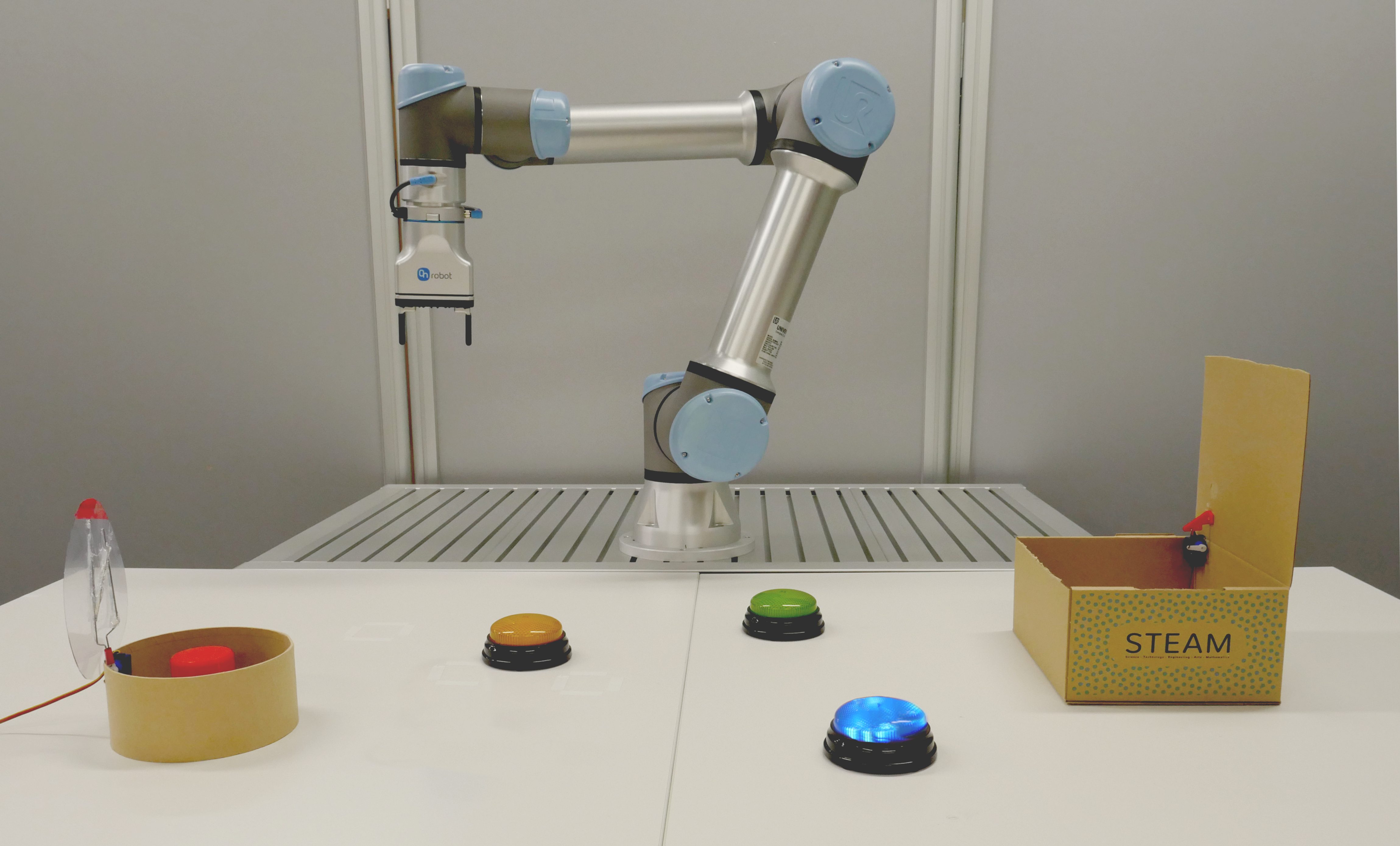}}
    \subfigure[]{\includegraphics[width=0.29\textwidth]{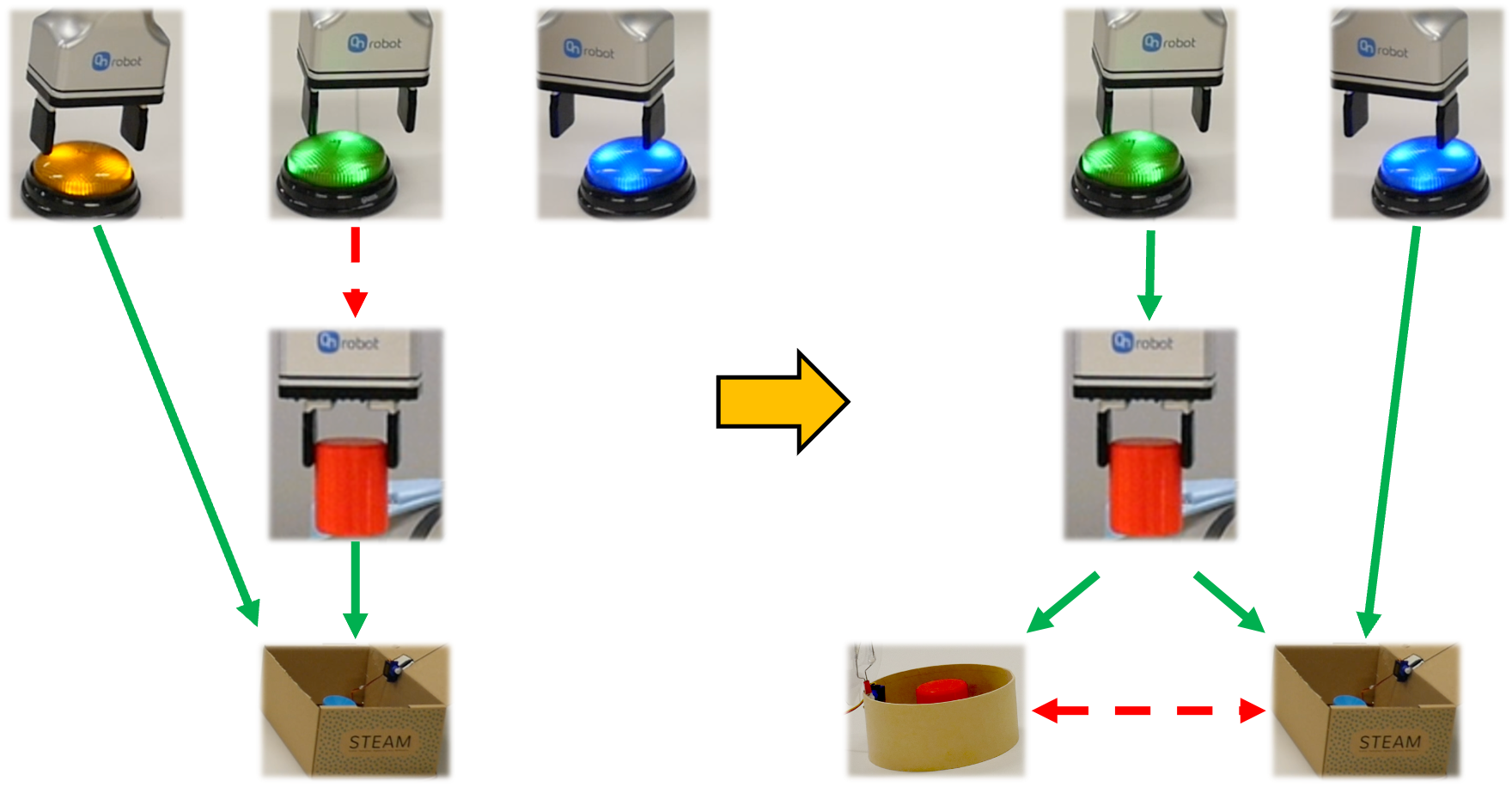}}
   \caption{(a) Experimental setup. (b) Interdependencies between objects in the tested scenario.}
    \label{Fig:setup}
\end{figure}

The scenario involves a UR5e robot (Fig. \ref{Fig:setup} (a)) and a table with cylinders, boxes and buttons that light up when pressed. In the experiments the experts control the direction of movement, the height of the arm, and its speed. The perception of the robot at each instant of time is a vector containing the relative distances between the objects and the robot end-effector ($d_j$), the states of the different buttons (pressed/not-pressed) ($s_i$), and the state of the force sensor in the gripper ($f$), indicating whether the gripper is occupied (gripping something) or not.

The robot must learn to perform multiple reach, pick and place tasks. A task here implies achieving a specific goal and, if necessary, the (sub-)goals that are preconditions for its achievement. The goals are related to what can be done with the different objects. Table \ref{tab:goals} summarises all the goals that can be discovered in this scenario. These goals also have dynamic interdependencies that evolve over time (see Fig. \ref{Fig:setup} (b)). Since goals are represented as events or changes in the robot's visual field, in this experimental setup, they are related to the interaction of the robotic arm with the different objects. When buttons are pressed, they light up, creating visual changes. When the robot grasps an object, the object will be hidden by the arm, causing it to ``disappear'' from the scene thus creating a visual change. Similarly, when a held object is released, it can ``reappear'' in a new position, producing another visual change.

\begin{table}
    \begin{center}
    \caption{Legend of discoverable goals. Once discovered, all of them can be selected either as goals or sub-goals.} 
    \label{tab:goals}
    \small
    \begin{tabular}{|p{2,5cm}|p{2,5cm}|p{2,5cm}|p{2,5cm}|p{2,5cm}|p{2,5cm}|p{2,5cm}|}
        \hline
        Goal 1 & Goal 2 & Goal 3 & Goal 4 & Goal 5 & Goal 6 \\ \hline
        Orange button pressed & Green button pressed & Blue button pressed & Red cylinder grabbed & Red cylinder in square box & Red cylinder in round box \\ \hline
    \end{tabular}
    \end{center}
\end{table}

\section{Experimental Results and Discussion} 
The experiment was run 10 times independently, each over 1500 epochs. An epoch ended upon reaching the goal or after 8 trials, while trials ended upon reaching the sub-goal or after 70 time steps. At each time step, the robot moved 5 cm. 

Initially, as shown in Fig. \ref{Fig:setup} (b), the environment included three buttons, a red cylinder inside a circular box, and a square box, with actions such as pressing buttons or retrieving the cylinder triggering specific outcomes. At epoch 750, the setup was modified: the orange button was removed, making its associated goal unachievable, and a new round box was introduced, adding a mutually exclusive goal with the square box (see Fig. \ref{Fig:setup} (b) after the yellow arrow).

\begin{figure}
    \centering
    \subfigure[]{\includegraphics[width=0.4\textwidth]{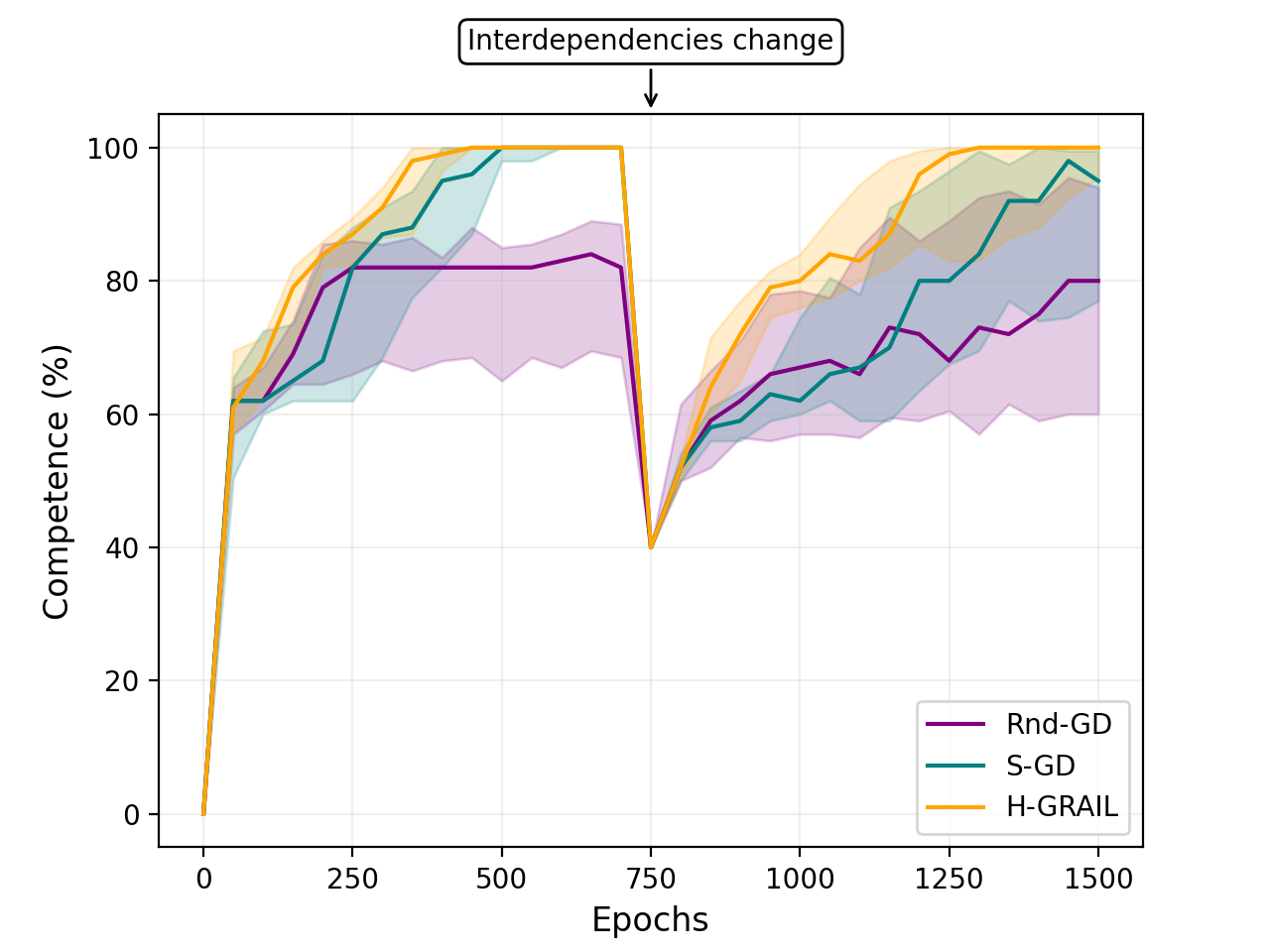}}
    \subfigure[]{\includegraphics[width=0.4\textwidth]{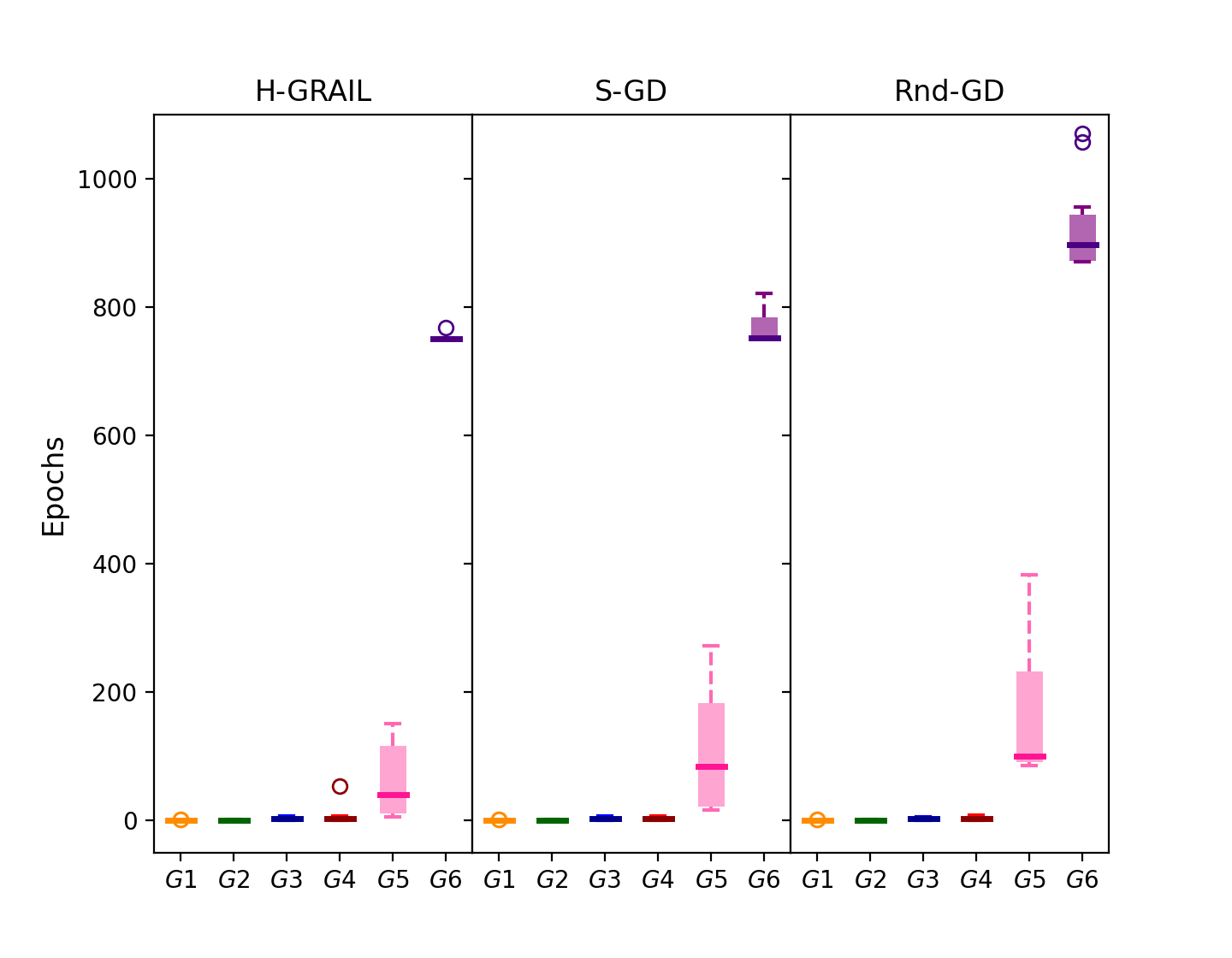}}
   \caption{(a) Robot competence over the set of discoverable goals for H-GRAIL and the two baselines. (b) Epochs of goal discovery for H-GRAIL and the two baselines Results derived from 10 independent runs of the experiment.}
    \label{Fig:results}
\end{figure}

The results depicted in Fig. \ref{Fig:results} (a) highlight H-GRAIL’s superior performance compared to the baseline systems. The performance metric indicates whether the robot learned the necessary skills and the curriculum (interdependencies between goals) to reach each of the goals within the 8 trials of each epoch. It can be seen that H-GRAIL not only discovered and learned goals more rapidly but also adapted more effectively to changes in goal interdependencies. In addition, Fig. \ref{Fig:results} (b) presents a boxplot showing the epochs at which the different goals were discovered. It highlights how H-GRAIL efficiently prioritized exploration, enabling the discovery of complex goals with preconditions (e.g., Goals 5 and 6) significantly earlier than the baselines. 

\bibliographystyle{abbrv}
\bibliography{RLDM2025_Ref}

\end{document}